\documentclass[11pt]{article}

\usepackage[preprint]{acl}
\usepackage{tikz}
\usepackage{multirow}

\usetikzlibrary{arrows.meta, positioning}

\usepackage{times}
\usepackage{latexsym}
\usepackage{amsmath} 
\usepackage{booktabs}
\usepackage{amsmath}
\DeclareMathOperator*{\argmax}{arg\,max}

\usepackage[T1]{fontenc}

\usepackage[utf8]{inputenc}

\usepackage{microtype}

\usepackage{inconsolata}

\usepackage{graphicx}

\title{Balancing Coverage and Draft Latency in Vocabulary Trimming for Faster Speculative Decoding}

\author{Ofir Ben Shoham \\
  Intuit \\
  \texttt{ofir\_benshoham@intuit.com} \\\
  }

\begin{document}
\maketitle
\begin{abstract}
Speculative decoding accelerates inference for Large Language Models by using a lightweight draft model to propose candidate tokens that are verified in parallel by a larger target model. Prior work shows that the draft model often dominates speculative decoding latency, since it generates tokens sequentially and incurs high cost from its language modeling head as vocabulary size grows. This exposes a fundamental trade-off in draft model design: larger vocabularies improve token coverage and agreement with the target model, but incur higher draft latency, while smaller vocabularies reduce latency at the risk of missing tokens required for accurate draft generation.

We address this trade-off through vocabulary trimming for draft models, motivated by the observation that domain-specific workloads use only a small fraction of the full vocabulary. We cast draft vocabulary selection as a constrained optimization problem that balances token coverage and draft latency. Coverage is computed over assistant responses in the training data, while latency is estimated using architecture-aware FLOPs that capture the cost of the language modeling head as a function of vocabulary size. We optimize a utility function with a Tree-structured Parzen Estimator to efficiently explore the coverage–latency Pareto frontier under a minimum coverage constraint. Experiments show improved speculative decoding throughput while reducing draft vocabularies by up to 97\% with high coverage. On domain-specific tasks, we achieve up to 16\% latency reduction and 20\% throughput improvement, and up to 6.7\% throughput gains on diverse out-of-distribution tasks.
\end{abstract}

\section{Introduction}
Applications of large language models (LLMs) have recently become widespread across many domains. However, Inference time remains a major bottleneck in the efficient deployment of large language models~\cite{leviathan2023fast}. Speculative Decoding has been proposed as a method to reduce inference latency by leveraging a smaller \emph{draft} model to generate candidate tokens, which are then verified by a larger model. Instead of performing $K$ separate forward passes, the large model verifies the proposed $K$ tokens in a single forward pass, significantly improving inference efficiency.

Recent work trains draft models on generations produced by the target model in order to improve the acceptance rate of draft tokens during verification by the large model, thereby reducing inference latency under speculative decoding~\cite{hong2025training}. However, draft models typically share the same vocabulary as the target model (e.g., 128K tokens for LLaMA~3), resulting in substantial computational overhead. This is particularly problematic given prior findings that draft model latency constitutes the primary bottleneck in speculative decoding \cite{yan2025decoding}. Moreover, \citet{goel2025vocabtrim} show that, for many downstream tasks, target model generation is confined to a small subset of the full vocabulary; for example, in function-calling tasks with LLaMA-3.2-3B-Instruct, more than 120K tokens are rarely or never generated.

To address these limitations and reduce draft model latency, recent work has explored vocabulary reduction techniques for the draft in speculative decoding. \citet{goel2025vocabtrim} propose VocabTrim, which trims infrequent tokens from the draft vocabulary based on token frequency statistics obtained from top-$p$ sampling during inference on a calibration dataset, increasing speculative decoding speedup by up to 16\% for LLaMA-3 models on Spec-Bench \citep{xia-etal-2024-unlocking}. Concurrently, \citet{zhao2025fr} introduce FR-Spec (Frequency-Ranked Speculative Sampling), which constrains the draft model's token selection to a frequency-ranked subset of the vocabulary, reducing LM head computation overhead by 75\% and achieving an average speedup of 1.12$\times$ over EAGLE-2 \citep{li2024eagle}.

However, selecting a fixed top-$k$ vocabulary can be suboptimal due to the inherent trade-off between draft model latency and vocabulary coverage. Larger vocabularies provide better coverage of the target model’s token distribution, but at the cost of increased draft inference latency. Furthermore, EAGLE-3~\cite{li2025eagle} embeds vocabulary mapping within the model weights via draft-to-target and target-to-draft index buffers, coupling vocabulary selection to training time. As a consequence, inference-time vocabulary reduction methods such as FR-Spec are incompatible with EAGLE-3, as noted in the SGLang inference engine documentation~\cite{zheng2024sglang}.

To address the limitations of prior work, we propose a novel approach that explicitly balances the trade-off between vocabulary coverage and draft model latency. Our approach formulates this trade-off through a reward function, which is optimized using the Tree-structured Parzen Estimator (TPE) \cite{watanabe2023tree}. The reward is defined by a utility function that combines token coverage measured on the training dataset with an architecture-aware estimation of draft model FLOPs.

We evaluate our approach on both out-of-distribution benchmarks and task-specific settings. On out-of-distribution benchmarks, it improves generation throughput by up to 6.7\% compared to full-vocabulary baselines, while on task-specific settings it achieves up to 19.6\% reductions in latency and throughput.

\paragraph{Our Contributions:}
\begin{enumerate}
    \item We formulate draft vocabulary selection to optimize speculative decoding inference as a constrained optimization problem, and propose a vocabulary trimming approach that leverages token frequency statistics together with an architecture-aware latency estimate to select an efficient draft vocabulary.
    \item We empirically show that the resulting draft models improve LLM generation throughput across both out-of-distribution and domain-specific tasks.
    \item We open source our implementation to support future research.
    \footnote{\url{https://github.com/Ofir408/balanced-coverage-latency-spec-decoding}}

\end{enumerate}

\section{Method}

In this work, we present a vocabulary trimming approach for speculative decoding that optimizes the trade-off between token coverage and draft model latency. Our method consists of five components: (1) formulating vocabulary selection as constrained optimization, (2) computing token coverage from training data, (3) estimating draft model FLOPs, (4) defining a utility function, and (5) optimizing via Tree-structured Parzen Estimator.

\subsection{Problem Formulation}

Let $\mathcal{V}$ denote the target model's vocabulary with $|\mathcal{V}| = V$ tokens. We seek a reduced draft vocabulary $\mathcal{V}_d \subset \mathcal{V}$ of size $k = |\mathcal{V}_d|$ that balances two competing objectives: maximizing token coverage while minimizing draft model latency.

We formulate this as a constrained optimization problem:
\begin{equation}
    k^* = \argmax_{k \in [k_{\min}, k_{\max}]} \; U(k) \quad \text{s.t.} \quad C(k) \geq c_{\min}
\end{equation}
where $U(k)$ is a utility function combining coverage and latency reduction, $C(k)$ is the token coverage at vocabulary size $k$, and $c_{\min}$ is a minimum coverage threshold.

The draft vocabulary $\mathcal{V}_d$ is constructed by selecting the top-$k$ most frequent tokens from the training distribution, ensuring that the most commonly generated tokens are retained.

\subsection{Token Coverage Estimation}

Following standard practice in instruction tuning, draft models are trained with a loss computed only over assistant response tokens, while user prompts and system messages are masked. We align our coverage metric with this training objective by counting token frequencies exclusively within assistant responses.

Given a training dataset of conversations, we parse assistant spans using chat template delimiters and compute token frequencies $f(v)$ for each token $v$ within these spans. Token coverage for a draft vocabulary of size $k$ is:
\begin{equation}
    C(k) = \frac{\sum_{v \in \text{top-}k} f(v)}{\sum_{v \in \mathcal{V}} f(v)}
\end{equation}
where top-$k$ denotes the $k$ most frequent tokens. This metric captures the fraction of training tokens covered by the draft vocabulary.


\subsection{Draft Model Latency Estimation}

We estimate draft model latency using FLOPs as a proxy. EAGLE-style draft models~\cite{li2025eagle} consist of three components:
\begin{enumerate}
    \item A \textbf{feature fusion layer} that projects concatenated hidden states from multiple target model layers into the hidden dimension.
    \item A single \textbf{transformer decoder layer} with self-attention and a feed-forward network.
    \item A \textbf{language modeling head} that projects hidden states to vocabulary logits.
\end{enumerate}

The key observation is that only the LM head depends on vocabulary size. For a linear projection from hidden dimension $d$ to vocabulary size $k$, the FLOPs are $2dk$. All other components contribute a fixed cost $F_{\text{fixed}}$ independent of $k$.

Table~\ref{tab:flops_breakdown} shows the FLOPs breakdown for LLaMA-3-8B. The LM head with full vocabulary ($V{=}128{,}256$) accounts for 64\% of total draft model computation, making vocabulary reduction highly effective.

\begin{table}[h]
\centering
\small
\begin{tabular}{lrr}
\toprule
Component & FLOPs & \% of Total \\
\midrule
Feature fusion & 100.7M & 6.1\% \\
Attention & 436.2M & 26.6\% \\
Feed-forward & 50.3M & 3.1\% \\
LM head ($V{=}128$K) & 1050.7M & 64.2\% \\
\midrule
Total & 1637.9M & 100\% \\
\bottomrule
\end{tabular}
\caption{Draft model FLOPs breakdown for LLaMA-3-8B.}
\label{tab:flops_breakdown}
\end{table}

We define the latency reduction as:
\begin{equation}
    R(k) = 1 - \frac{F_{\text{fixed}} + 2dk}{F_{\text{fixed}} + 2dV}
\end{equation}


\subsection{Utility Function Definition}

We define a utility function that represents that trade-off between token coverage and latency reduction with a tunable weight parameter $\alpha \in [0,1]$:
\begin{equation}
    U(k) = \alpha \cdot C(k) + (1 - \alpha) \cdot R(k)
\end{equation}

The coverage weight $\alpha$ controls the trade-off between maintaining high token coverage (larger $\alpha$) and maximizing draft's FLOPs latency reduction (smaller $\alpha$). Higher $\alpha$ values prioritize draft accuracy at the cost of computational savings, while lower values aggressively reduce vocabulary size.

The utility function formulation of the trade-off between token coverage and draft model latency enables exploration of the Pareto frontier between target-vocabulary token coverage and draft model latency, allowing TPE optimization of this objective to select the most accurate draft vocabulary size.

\subsection{TPE-Based Optimization}

We optimize the vocabulary size using TPE, a sequential model-based optimization algorithm well-suited for hyperparameter search.

TPE models the objective function by maintaining two density estimators: $\ell(k)$ for vocabulary sizes that yield high utility, and $g(k)$ for those with low utility. At each iteration, TPE samples candidate vocabulary sizes by maximizing the ratio $\ell(k)/g(k)$, which serves as a proxy for expected improvement.

To enforce the minimum coverage constraint $C(k) \ge c_{\text{min}}$, we modify the objective to return a penalty value when the constraint is violated:
\begin{equation}
\tilde{U}(k) =
\begin{cases}
U(k) & \text{if } C(k) \ge c_{\text{min}} \\
-1   & \text{otherwise}.
\end{cases}
\end{equation}

This formulation guides TPE to learn the feasible region while still exploring the coverage-latency trade-off within that region. The optimization runs for $N$ trials (we use $N=100$), with TPE progressively focusing on promising regions of the search space.

The output is the optimal vocabulary size $k^*$ that maximizes utility while satisfying the coverage constraint. The corresponding draft vocabulary $\mathcal{V}_d$ consists of the $k^*$ most frequent tokens from the training distribution.

\subsection{Draft Vocabulary Size Optimization}

\begin{figure}[t]
\centering
\includegraphics[width=0.8\columnwidth]{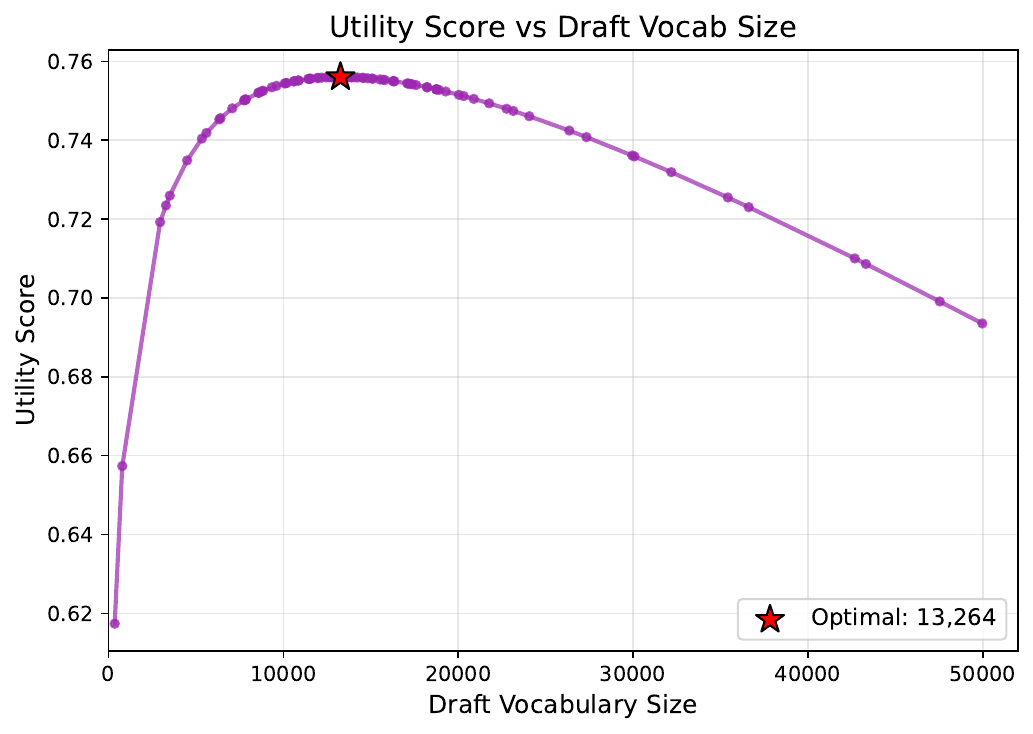}
\caption{Utility score vs.\ draft vocabulary size. The utility function peaks at 
13,264 tokens, representing the optimal tradeoff between coverage and latency while training on the Open-PerfectBlend dataset \cite{xu2024perfect}.}
\label{fig:utility}
\end{figure}

Figure~\ref{fig:utility} shows the utility score across 100 vocabulary configurations 
explored via Optuna's TPE sampler \cite{akiba2019optuna}. The curve exhibits a clear maximum at 13,264 tokens, demonstrating the optimal balance point. Smaller vocabularies suffer from insufficient  coverage despite greater speedup, while draft larger vocabularies provide diminishing coverage  gains that do not justify their draft latency cost.

\begin{figure}[t]
\centering
\includegraphics[width=0.85\columnwidth]{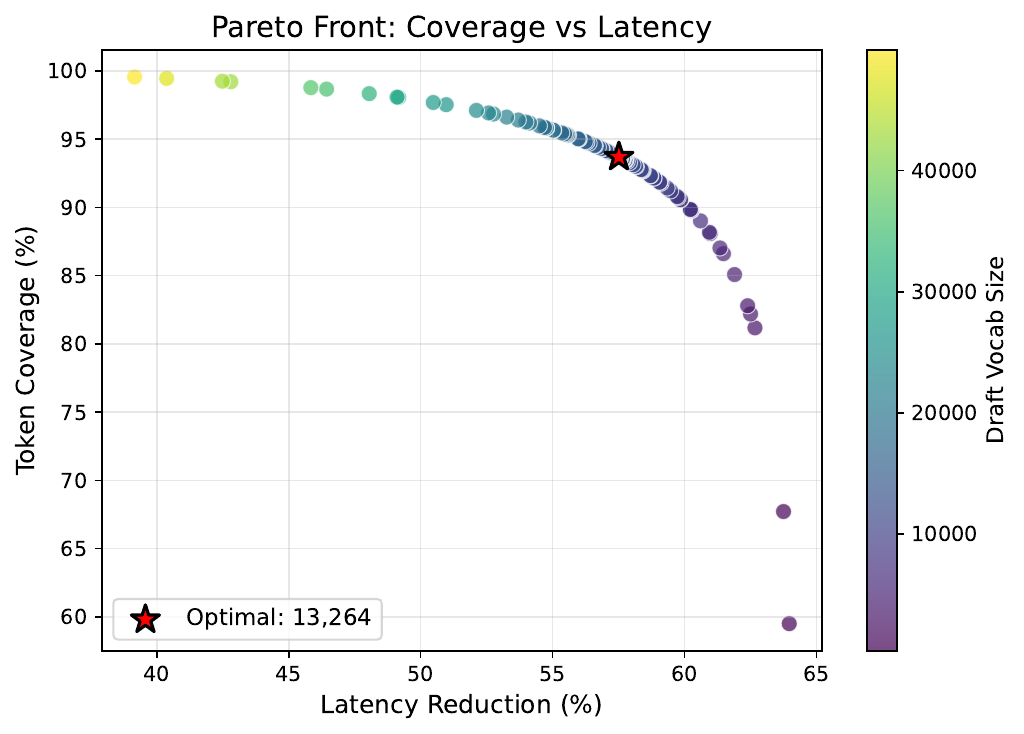}
\caption{Pareto front showing the coverage-latency tradeoff of the draft model, while training on the Open-PerfectBlend dataset \cite{xu2024perfect}. The optimal point 
(red star) balances high coverage with substantial latency reduction.}
\label{fig:pareto}
\end{figure}

Figure~\ref{fig:pareto} visualizes the Pareto frontier of the coverage-latency tradeoff. 
Each point represents a vocabulary configuration (color-coded by size), revealing 
the fundamental tension: smaller vocabularies achieve up to 64\% latency reduction 
but cover only 60\% of tokens, while near-complete coverage (99\%+) limits latency 
reduction to 39\%. The optimal configuration achieves 93.7\% coverage with 57.5\% 
LM head latency reduction (a 90\% vocabulary reduction 128K$\rightarrow$13K tokens).

\section{Experiments}
To evaluate our approach, we adopt Llama-3.1-8B-Instruct \cite{grattafiori2024llama} as the target model and use datasets to regenerate responses with Llama-3.1-8B-Instruct, ensuring alignment between the draft model and the target model it is trained to accelerate.

We then measure the inference throughput of the resulting draft model using the SpecForge framework \cite{specforge2025} together with the SGLang inference engine \cite{zheng2024sglang}. All experiments are conducted with 3 independent runs, and we report mean values with 95\% confidence intervals to ensure statistical rigor. Evaluations are conducted on benchmarks that are out of distribution relative to the training data, including MT-Bench (multi-turn conversational and instruction-following quality) \cite{zheng2023judging}, GSM8K (grade-school mathematical reasoning) \cite{cobbe2021training}, HumanEval (code generation and functional correctness) \cite{chen2021evaluating}, MATH500 (advanced mathematical problem solving) \cite{lightman2023lets}, and AIME (competition-level mathematical reasoning) \cite{aime25}. We also measure our approach on two internal downstream tasks, including Named Entity Recognition and function calling.

\subsection{Results on out of distribution datasets}
We train two draft models on the Open-PerfectBlend dataset~\cite{xu2024perfect}: one using a 13264 tokens vocabulary derived from our approach, and one using the full 128K LLaMA-3 vocabulary as a baseline. Both models are evaluated on out-of-distribution benchmarks to assess generalization. As shown in Table~\ref{tab:throughput_diff}, the vocabulary-trimmed draft model consistently outperforms the baseline, with throughput improvements ranging from 2.2\% to 6.7\%.


Despite reducing vocabulary size by approximately 90\%, the draft model maintains effective acceleration across all out-of-distribution benchmarks, demonstrating that the coverage learned from Open-PerfectBlend generalizes beyond the training distribution.

\begin{table}[h]
\centering
\small
\begin{tabular}{lccc}
\hline
\textbf{Benchmark} & \textbf{Our approach} & \textbf{128k-vocab} & \textbf{Diff (\%)} \\
\hline
MT-Bench   & \textbf{177.54 $\pm$ 0.52} & 172.38 $\pm$ 0.67  & +3.0\% \\
GSM8K      & \textbf{160.51 $\pm$ 0.50} & 155.47 $\pm$ 0.43  & +3.2\% \\
HumanEval  & \textbf{206.79 $\pm$ 0.62} & 202.31 $\pm$ 0.72  & +2.2\% \\
MATH500    & \textbf{217.22 $\pm$ 0.40} & 206.74 $\pm$ 1.14  & +5.1\% \\
AIME       & \textbf{224.83 $\pm$ 1.30} & 210.74 $\pm$ 0.69  & +6.7\% \\
\hline
\end{tabular}
\caption{Output throughput comparison between our approach and a baseline with the full vocabulary, on a setup of a single A100-80G GPU. Values shown as mean $\pm$ 95\% CI over 3 runs. Higher is better.}
\label{tab:throughput_diff}
\end{table}

\subsection{In-domain Results}
Beyond general-purpose benchmarks, we validate our approach on two domain-specific tasks: Named Entity Recognition (NER) and Function Calling. For each task, we fine-tune a target model and train two different draft models, one with our optimized vocabulary and one with the full 128K LLaMA-3 vocabulary. Our approach yields vocabularies of just 6{,}521 tokens for NER and 4{,}380 tokens for Function Calling, representing 95\% and 97\% reductions respectively.

\begin{table*}[t]
\centering
\caption{
In-domain benchmark results on a single NVIDIA A100-80GB GPU over 500 samples.
For each task, percentage deltas are reported relative to the task-specific baseline.
Values shown as mean $\pm$ 95\% CI over 3 runs.
Lower latency is better; higher output throughput is better.
}

\label{tab:in_domain_results}
\begin{tabular}{llcccc}
\hline
\textbf{Benchmark} &
\textbf{Model / Vocabulary} &
\textbf{Latency (s)} &
$\Delta$ (\%) &
\textbf{Output Throughput} &
$\Delta$ (\%) \\
\hline
\multirow{2}{*}{NER}
& 128K vocab (baseline)
& 468.11 $\pm$ 5.50 & -- & 146.50 $\pm$ 1.72 & -- \\

& 6{,}521 vocab (Our Approach)
& \textbf{391.43 $\pm$ 3.33} & \textbf{-16.4\%}
& \textbf{175.18 $\pm$ 1.49} & \textbf{+19.6\%} \\
\hline
\multirow{2}{*}{Function Calling}
& 128K vocab (baseline)
& 401.02 $\pm$ 4.28 & -- & 192.90 $\pm$ 2.06 & -- \\

& 4{,}380 vocab (Our Approach)
& \textbf{364.54 $\pm$ 4.36} & \textbf{-9.1\%}
& \textbf{212.21 $\pm$ 2.56} & \textbf{+10.0\%} \\
\hline
\end{tabular}
\end{table*}

Table~\ref{tab:in_domain_results} reports results on a single NVIDIA A100-80GB GPU. The optimized draft models consistently outperform the 128K-vocabulary baselines: the NER task shows 16.4\% latency reduction and 19.6\% throughput improvement, while Function Calling achieves 9.1\% latency reduction and 10.0\% throughput improvement. These results confirm that task-aligned vocabulary optimization yields substantial efficiency improvements for speculative decoding in domain-specific applications.

\section{Discussion}
\subsection{Generalization to Out-of-Distribution Datasets}
\label{sec:generalization}

The out-of-distribution results in Table~\ref{tab:throughput_diff} show consistent throughput improvements across diverse benchmarks, despite the vocabulary being optimized on a different dataset. To understand why vocabulary trimming generalizes effectively, we analyze the token coverage of our 13K optimized vocabulary on actual target model generations across the five OOD benchmarks.

\paragraph{Token Coverage Analysis.}
Table~\ref{tab:ood-coverage} presents the frequency-weighted token coverage achieved by our optimized vocabulary on target model outputs for each benchmark. Despite reducing vocabulary size by approximately 90\% (from 128K to 13K tokens), the optimized vocabulary achieves 97.1\% average coverage across all OOD benchmark generations, with coverage ranging from 93.2\% on MT-Bench to 98.6\% on MATH-500 and AIME.

\begin{table}[h]
\centering
\small
\caption{Token coverage of the 13K optimized vocabulary on target model generations for out-of-distribution benchmarks. Freq.\ Coverage measures the fraction of generated token occurrences covered; Unique Coverage measures the fraction of distinct generated tokens covered.}
\label{tab:ood-coverage}
\begin{tabular}{lcccc}
\toprule
Benchmark & Tokens & Freq.\ Cov. & Unique Cov. \\
\midrule
GSM8K & 117,613 & 94.6\% & 69.0\% \\
MT-Bench & 59,439 & 93.2\% & 70.4\% \\
HumanEval & 61,056 & 97.0\% & 75.6\% \\
MATH-500 & 295,210 & 98.6\% & 78.3\% \\
AIME & 40,850 & 98.6\% & 87.3\% \\
\midrule
\textbf{Combined} & 574,168 & \textbf{97.1\%} & --- \\
\bottomrule
\end{tabular}
\end{table}

The gap between frequency-weighted coverage (97.1\%) and unique token coverage reveals a key insight: while many unique tokens fall outside the optimized vocabulary, these tokens appear infrequently. The missing tokens are predominantly task-specific terminology that appears rarely in generation outputs.

\paragraph{Domain Alignment with Training Data.}
The OpenPerfectBlend training dataset comprises a diverse mixture of instruction-tuning data across multiple domains: mathematics (39.4\%), code (38.9\%), chat (17.6\%), and instruction following (4.1\%) \cite{xu2024perfect}. Table~\ref{tab:domain-coverage} shows how benchmark coverage correlates with training data domain representation.

\begin{table}[h]
\centering
\small
\caption{OOD benchmark coverage by training data domain.}
\label{tab:domain-coverage}
\begin{tabular}{llcc}
\toprule
Benchmark & Domain & Coverage & Training \% \\
\midrule
GSM8K & Math & 94.6\% & 39.4\% \\
MATH-500 & Math & 98.6\% & 39.4\% \\
AIME & Math & 98.6\% & 39.4\% \\
HumanEval & Code & 97.0\% & 38.9\% \\
MT-Bench & Chat & 93.2\% & 17.6\% \\
\bottomrule
\end{tabular}
\end{table}

Notably, the training and evaluation datasets differ substantially in their construction and difficulty levels. The math training data includes MetaMathQA, which bootstraps questions from GSM8K and MATH training sets by ``rewriting the question from multiple perspectives without extra knowledge'' \cite{yu2024metamath}. Evaluation is performed on the GSM8K and MATH-500 \emph{test sets}, which contain unseen problems, as well as AIME (olympiad-level), with the latter presenting substantially harder problems than the training distribution. Similarly, code training uses evol-codealpaca \cite{luo2023wizardcoder} and UltraInteract \cite{yuan2024advancing}, while evaluation uses HumanEval with its distinct function-completion format and doctest syntax. Despite these differences in problem difficulty and format, the vocabulary achieves 94--99\% coverage on model generations, demonstrating that core mathematical notation, operators, and common programming tokens are shared across difficulty levels. Even MT-Bench, representing the chat domain with only 17.6\% of training data, achieves 93.2\% coverage, indicating that our approach captures essential high-frequency tokens even for underrepresented domains.

\paragraph{Missing Token Analysis.}
Examining the tokens excluded from the optimized vocabulary reveals that missing tokens are predominantly task-specific and low-frequency. GSM8K's lower coverage (94.6\%) is primarily due to the \texttt{<<} annotation token used in GSM8K's answer format (appearing 3,275 times), along with domain-specific nouns from word problem narratives (e.g., ``mashed,'' ``fries,'' ``laundry'').
HumanEval misses Python doctest syntax (\texttt{>>>}) and code-specific tokens (\texttt{\_shift}, \texttt{enumerate}).
MATH-500 and AIME exclude mathematical terminology (``asympt,'' ``cyclic,'' ``diamond'') and geometry-related tokens used in competition problems.
MT-Bench lacks proper nouns (``Wars,'' ``Britain,'' ``Socrates'') and rare conversational vocabulary (``seismic,'' ``settlement'').
Critically, these missing tokens appear infrequently enough that their absence minimally impacts the draft model's ability to propose acceptable candidates.

Taken together, these findings explain why vocabulary trimming generalizes effectively: high-frequency tokens are largely domain-agnostic and captured by diverse training data, while the long tail of task-specific tokens can be safely excluded without degrading speculative decoding performance. The balanced composition of OpenPerfectBlend spanning math, code, and chat domains, ensures that the optimized vocabulary covers the essential tokens needed across diverse tasks, enabling draft vocabulary optimization to generalize to out-of-distribution benchmarks with 97\% coverage on actual model generations.

\subsection{Accept Length and Throughput Analysis}
\label{sec:accept-length-throughput-analysis}

Table~\ref{tab:downstream-analysis} presents a comprehensive comparison of vocabulary trimming effects across out-of-distribution benchmarks and downstream tasks, relating draft vocabulary size, token coverage, accept length, and throughput improvement.

\begin{table}[h]
\centering
\tiny
\caption{Analysis of vocabulary size, token coverage, accept length, and throughput improvement across OOD benchmarks and downstream tasks.}
\label{tab:downstream-analysis}
\begin{tabular}{lcccccc}
\toprule
\textbf{Task} & \textbf{Draft Vocab} & \textbf{Coverage} & \multicolumn{2}{c}{\textbf{Accept Length}} & \textbf{Throughput $\Delta$} \\
 & & & 128K & Ours & \\
\midrule
\multicolumn{6}{l}{\textit{Out-of-Distribution Benchmarks}} \\
MT-Bench & 13,264 & 93.2\% & 2.89 & 2.65 & +3.0\% \\
GSM8K & 13,264 & 94.6\% & 2.75 & 2.53 & +3.2\% \\
HumanEval & 13,264 & 97.0\% & 3.39 & 3.09 & +2.2\% \\
MATH-500 & 13,264 & 98.6\% & 3.45 & 3.23 & +5.1\% \\
AIME & 13,264 & 98.6\% & 3.53 & 3.35 & +6.7\% \\
\midrule
\multicolumn{6}{l}{\textit{Downstream Tasks}} \\
Function Calling & 4,380 & 98.6\% & 3.86 & 3.82 & +10.0\% \\
NER & 6,521 & 85.1\% & 1.69 & 1.69 & +19.6\% \\
\bottomrule
\end{tabular}
\end{table}

The results demonstrate that vocabulary trimming improves throughput even when accept length decreases, because the reduction in draft model latency more than compensates for the decrease in acceptance rate. The draft model executes faster per speculation cycle, and this speedup accumulates across all cycles to yield net throughput gains. This effect is observed across both out-of-distribution benchmarks and downstream tasks, though with different magnitudes.

For out-of-distribution benchmarks, accept length reduction ranges from 5--9\%, which is expected since these benchmarks evaluate on tasks not seen during vocabulary optimization. The optimized vocabulary is therefore less aligned with the token distribution of these benchmarks. Nevertheless, throughput consistently improves by 2.2--6.7\% across all OOD benchmarks, confirming that the draft latency reduction outweighs the acceptance penalty.

Downstream tasks exhibit even smaller accept length degradation due to domain-specific vocabulary optimization. In the function-calling setting, the accept length is reduced by only $1.0\%$ (from $3.86$ to $3.82$) when shrinking the vocabulary from $128\text{K}$ to $4{,}380$ tokens, while throughput increases by $10.0\%$. NER exhibits an even more favorable trade-off: despite achieving only 85.1\% token coverage, the accept length remains unchanged (1.69$\rightarrow$1.69), resulting in a 19.6\% throughput improvement. The higher improvement compared to function calling can be attributed to the complete preservation of accept length, allowing the full benefit of vocabulary reduction to translate directly into throughput gains without any acceptance penalty. These results demonstrate that domain-specific vocabulary optimization can achieve larger throughput gains with smaller vocabularies compared to general-purpose optimization.

\subsection{Stability of Vocabulary Optimization}
\begin{figure}[h]
\centering
\includegraphics[width=0.85\columnwidth]{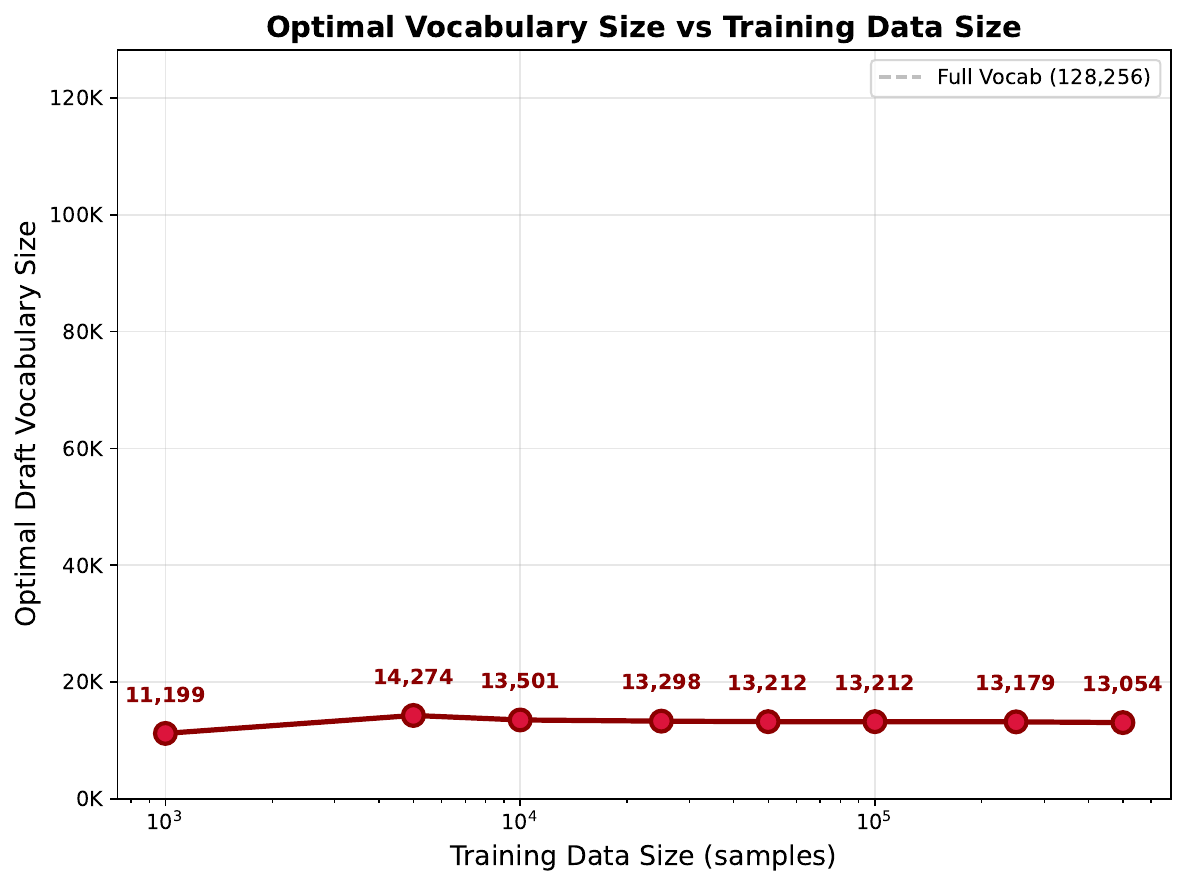}
\caption{Optimal vocabulary size based on our approach vs.\ training data size. The optimal vocabulary converges to approximately 13K tokens after 10K samples, demonstrating stability across different random subsets of the training data.}
\label{fig:vocab-stability}
\end{figure}

To verify that our optimization approach produces stable results regardless of training data sampling, we analyze how the optimal vocabulary size varies with different amounts of training data. Figure~\ref{fig:vocab-stability} shows the optimal vocabulary size identified by our approach when using subsets of the OpenPerfectBlend dataset ranging from 1,000 to 500,000 randomly selected samples.

The optimal vocabulary size converges rapidly, stabilizing around 13,000--13,300 tokens after approximately 10,000 training samples. Beyond this point, increasing the training data from 25K to 500K samples (a 20$\times$ increase) changes the optimal vocabulary by less than 2\% (from 13,298 to 13,054 tokens). This stability indicates that the token frequency distribution captured by our approach is robust to random sampling variation. Even with only 1,000 samples, the optimization identifies a vocabulary (11,199 tokens) within 15\% of the converged value, suggesting that the high-frequency tokens critical for coverage are consistently represented across random subsets. These results demonstrate that practitioners can reliably optimize vocabulary size using moderate-sized subsets of their training data, without requiring access to the full dataset.

\section{Limitations}

Our approach has several limitations. First, we evaluate only on LLaMA-3.1-8B-Instruct as the target model; generalization to other model families (e.g., Qwen, Gemma, Mistral) and larger model scales (70B, 405B) remains to be validated.

Second, our approach requires training the draft model with the reduced vocabulary, unlike inference-time methods such as VocabTrim~\cite{goel2025vocabtrim} that prune the vocabulary post-training for Eagle2. While training-time reduction avoids distribution mismatch between training and inference, it incurs additional computational cost when adapting to new domains and prevents retrofitting existing draft models without retraining.

Third, we evaluate exclusively on EAGLE3~\cite{li2025eagle} speculative decoding within the SGLang inference engine. Future work should validate vocabulary trimming across a broader range of speculative decoding frameworks.

\section{Conclusions}
We introduced vocabulary trimming for speculative decoding draft models, targeting the vocabulary-dependent cost of the LM head, a major contributor to draft latency. We formulated draft vocabulary selection as a constrained optimization problem that balances frequency-weighted token coverage (computed over assistant responses) with an architecture-aware FLOPs proxy for latency, and used TPE to efficiently search the coverage--latency trade-off.

Empirically, trimmed-vocabulary drafts improve end-to-end speculative decoding throughput across both general and domain-specific settings. A 13{,}264-token draft vocabulary (about a 90\% reduction from 128K) yields consistent throughput gains on out-of-distribution benchmarks, up to \(+6.7\%\), while maintaining high frequency-weighted coverage on target generations. In-domain optimization enables more aggressive trimming (4.4K--6.5K tokens) and delivers larger improvements, reaching 19.6\% and 10.0\% throughput improvements. These results show that optimizing draft vocabulary size is a simple, robust mechanism for accelerating speculative decoding without sacrificing practical coverage, especially when aligned to the deployment domain.

\bibliography{custom}




\end{document}